%% file: neurips_2025.tex
\documentclass{article}

\usepackage{comment} 

 \usepackage[preprint]{neurips_2025}

\usepackage[utf8]{inputenc} 
\usepackage[T1]{fontenc}    
\usepackage{hyperref}       
\usepackage{url}            
\usepackage{booktabs}       
\usepackage{amsfonts}       
\usepackage{nicefrac}       
\usepackage{microtype}      
\usepackage{xcolor}         
\usepackage{wrapfig}        
\usepackage{subcaption}

\usepackage{graphicx}
\usepackage{tabularx}

\usepackage{booktabs}
\usepackage{supertabular}
\usepackage{longtable}  
\usepackage{subcaption}
\usepackage{longtable}

\usepackage{tikz}
\usetikzlibrary{trees}
\usepackage{enumitem,amssymb}

\input{commands}

\title{Evaluation of Cultural Competence of Vision-Language Models}

\author{
\textbf{Srishti Yadav}\textsuperscript{1,5,6}\thanks{Corresponding author: srya@di.ku.dk} , 
\textbf{Lauren Tilton}\textsuperscript{2},
\textbf{Maria Antoniak}\textsuperscript{1,6},
\textbf{Taylor Arnold}\textsuperscript{3}, \\
\textbf{Jiaang Li}\textsuperscript{1,6}, 
\textbf{Siddhesh Milind Pawar}\textsuperscript{1},
\textbf{Antonia Karamolegkou}\textsuperscript{1},
\textbf{Stella Frank}\textsuperscript{1,6},
\textbf{Zhaochong An}\textsuperscript{1,6}, \\
\textbf{Negar Rostamzadeh}\textsuperscript{4},    
\textbf{Daniel Hershcovich}\textsuperscript{1},    
\textbf{Serge Belongie}\textsuperscript{1,6},
\textbf{Ekaterina Shutova}\textsuperscript{5} \\[0.2cm]
\textsuperscript{1}Department of Computer Science, University of Copenhagen, Denmark \\
\textsuperscript{2}Department of Rhetoric and Communication Studies, University of Richmond, U.S.A. \\
\textsuperscript{3}Department of Data Science and Statistics, University of Richmond, U.S.A. \\
\textsuperscript{4}Google Research \\
\textsuperscript{5}ILLC, University of Amsterdam, Netherlands \\
\textsuperscript{6}Pioneer Centre of AI, Denmark 
}

\begin{document}

\maketitle

\begin{abstract}

Modern vision-language models (VLMs) often fail at cultural competency evaluations and benchmarks. Given the diversity of applications built upon VLMs, there is renewed interest in understanding how they encode cultural nuances. While individual aspects of this problem have been studied, we still lack a comprehensive framework for systematically identifying and annotating the nuanced cultural dimensions present in images for VLMs. This position paper argues that foundational methodologies from visual culture studies (cultural studies, semiotics, and visual studies) are necessary for cultural analysis of images. Building upon this review, we propose a set of five frameworks, corresponding to cultural dimensions, that must be considered for a more complete analysis of the cultural competencies of VLMs. 
\end{abstract}

\section{Introduction}

Studying cultural competencies of large AI models is essential to prevent the amplification of cultural biases, ensure fair representation, and promote more inclusive, context-sensitive ML systems \citep{bhatt-diaz-2024-extrinsic}. 
In natural language processing (NLP), there has been a flurry of recent work measuring the socio-cultural dimensions of large language models (LLMs), including how LLMs encode, express, and respond to culturally situated prompts \citep{hershcovich2022challenges, liu2024culturally, zhou2025culture}. 
For instance, recent studies have examined value alignment \citep{choenni2024self}, moral reasoning across languages \citep{agarwal2024ethical}, and cultural personas \citep{alkhamissi2024investigating}, while also uncovering strong Western biases in model outputs \citep{jha2024visage}, which can threaten cultural diversity if deployed to users~\citep{bird-2020-decolonising,bird-2024-must}. 
There have also been efforts to address these concerns, including prompting based on ethnographic fieldwork \citep{alkhamissi2024investigating} and fine-tuning culture-specific LLMs \citep{li2024culturellm}. For vision-language models (VLMs), there have been recent efforts to understand cultural representation through curated datasets \citep{liu2021visually, yin2021broaden} and evaluation of cultural awareness \citep{yadav2025beyond}.

\begin{figure}
  \centering
    \includegraphics[width=1.0\linewidth]{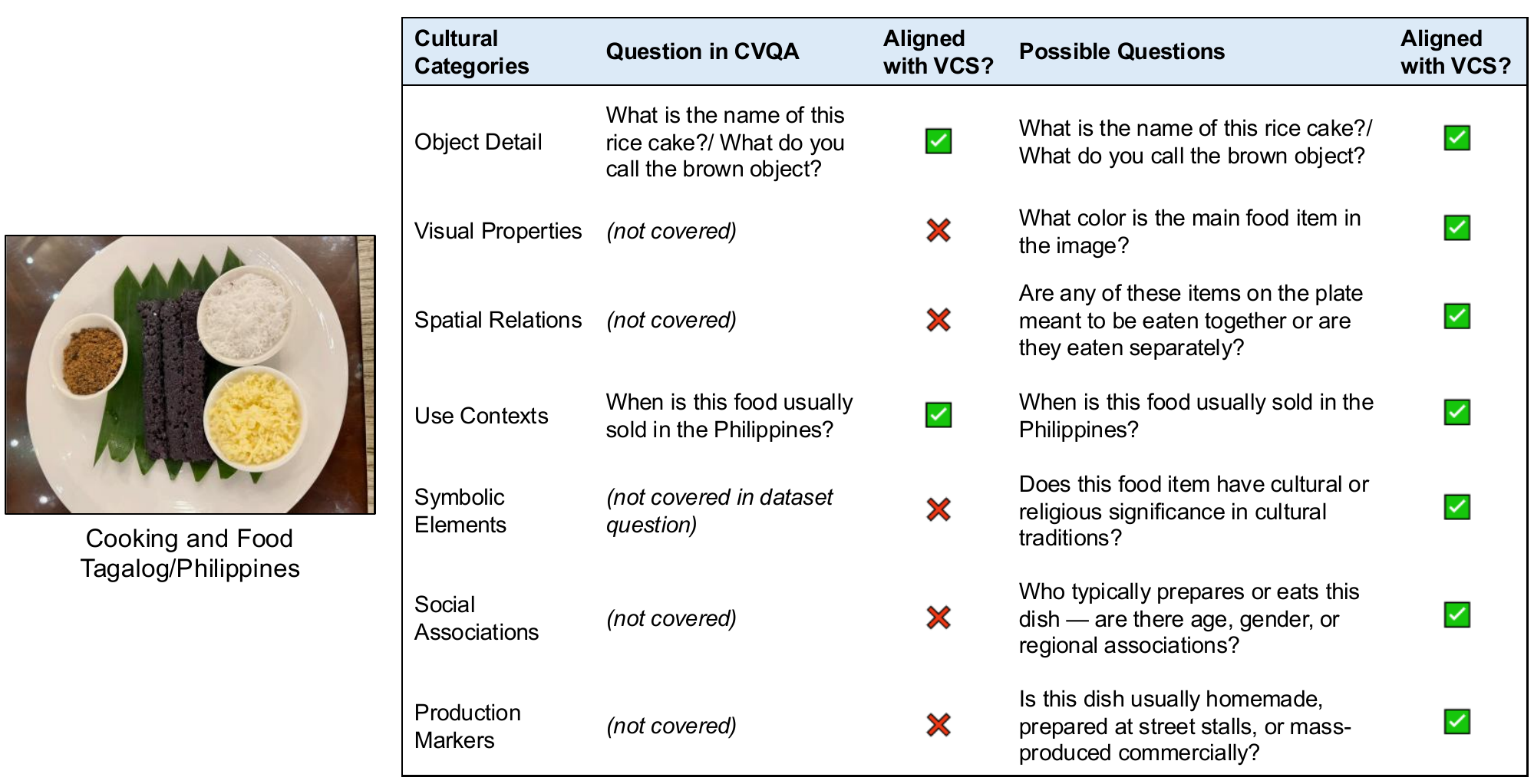}
    \caption{An image from a cultural evaluation dataset (CVQA \citep{romero2025cvqa}), compared against how  possible questions arising from Visual Cultural Studies (VCS). This gap exists in various datasets with the stated goal of studying cultural knowledge in VLMs, hence the need for frameworks grounded in visual cultural studies. }
    \label{fig:main-plot}
\end{figure}

Given the significant recent advances in multimodal\footnote{We distinguish between multimodal models, which could incorporate audio or other modalities, and VLMs, which incorporate images and text. In this paper, we focus on VLMs.} LLMs, specifically vision and language models, the field is now presented with
an urgent challenge: \textbf{the need for robust frameworks to evaluate multimodal cultural awareness that account for both visual and textual domains}.

We argue that such evaluations of VLMs' cultural competencies need to be informed by insights from three key fields: cultural studies, semiotics, and visual studies. Cultural studies is a capacious area of theory that brings together fields such as anthropology, history, media studies, and postcolonial theory to study cultural formations, including the values, beliefs, and ideas of communities as cultural forms, including images \citep{williams1983culture, hall1997representation}. Semiotics is the study of signs and sign systems - how meaning is created and communicated through symbols, words, images, sounds, and other forms. When put together, they ask us to take into account the ways in which specific forms, such as images, make meaning while accounting for the cultural, historical, and social specificity of how the form makes meaning and is interpreted \citep{peirce1868categories,saussure1916cours,barthes1977image}.
Finally, visual studies emerged at the intersection of these two fields to provide analytical specificity to how meaning is created and communicated through visual forms across time, place, and form \citep{berger1973ways, sturken2001practices, mitchell1995picture, mirzoeff1999introduction}. We will refer to these three interconnected fields as \textit{visual cultural studies} throughout the paper. 

This paper takes an interdisciplinary perspective to identify which dimensions of images have been studied across these fields and how they are tied to the visual understanding of cultural knowledge. We draw on these insights to put forward a proposal for more comprehensive frameworks to study cultural awareness of VLMs. Where possible, we also relate our proposals to current methods of evaluation. We also identify future possibilities for designing culturally-aware benchmarks, interpretability tools, and evaluation protocols for VLMs. We note that this paper does not aim to experiment with these frameworks --- we leave this for our research community to operationalise --- but rather to expand awareness of the set of dimensions of culture that are evident in images and provide an inventory of tools to study them in VLMs. \textbf{This position paper argues that we need a paradigm shift in developing and evaluating VLMs for better cultural representation: from surface-level cultural cues to deeper, theory-informed analysis rooted in decades of visual cultural research.} 
\par

\textbf{Contributions:} 
This position paper addresses three key questions on how to identify, interpret, and evaluate cultural knowledge in images using VLMs. 
In \S\ref{sec:multimodel_models}, we identify the gaps in the current literature on VLMs and culture. 
 In \S\ref{sec:visual_studies}, we survey research in visual cultural studies (cultural studies, semiotics, visual studies) that have studied images and how they reflect culture.
Using knowledge from visual cultural studies, we propose a set of five frameworks in~\S\ref{frameworks} to fill the gaps we found in \S\ref{sec:multimodel_models}.  
To the best of our knowledge, this is the first work bringing visual cultural studies to the attention of the ML community for grounding their current evaluations.

\section{VLMs and cultural awareness}
\label{sec:multimodel_models}


In the humanities, culture has long been a subject of discussions \citep{halpern1955dynamic,harris1968rise,Nisbett2003}. It is broadly understood as the learned behaviors and symbolic systems that shape how people live together in societies. This includes both tangible and intangible elements—language, beliefs, customs, institutions, tools, artistic practices, rituals, and ceremonies\footnote{https://www.britannica.com/dictionary/culture}. To adequately and equitably serve diverse human cultures, models (VLMs and LLMs) must be culturally aware. We want models that accurately interpret culturally specific visual and associated linguistic cues, generate culturally appropriate text related to visual input, and mitigate biases inherent in training data that disproportionately misrepresent certain cultures. 
 These goals are driven by the need for inclusive ML systems that are robust when deployed in our multicultural world. 
 To address these issues in a multimodal setting, recent works have proposed culturally-aware benchmarks and evaluation methods (details in \S\ref{sec:dataset-curation}).

\subsection{Dataset curation}
\label{sec:dataset-curation}


A standard way of evaluating the cultural knowledge of VLMs has been to curate datasets consisting of images of objects, entities, or situations corresponding to cultural \textit{concepts} \citep{murphy2004big}, together with human annotations. These datasets are then used to assess model performance on downstream tasks such as cultural visual question answering (VQA), cultural entity recognition, or image captioning. While these efforts provide a valuable starting point, they often reduce culture to its most coarse layer: naming or recognising entities within broad categories such as food, clothing, rituals,~etc. However, object labelling alone fails to reflect the fine-grained, situated, and symbolic nature of cultural understanding in human cognition.


The dataset curation process also varies from dataset to dataset, and more critically, most of the datasets rarely address deeper dimensions such as whether an item is used in domestic, communal, or ritual contexts; whether it conveys symbolic, historical or social meaning; or whether its form, materiality, or aesthetics carries culturally specific associations. 
These datasets cme from different sources. For example, datasets like CCSK \citep{nguyen2023extracting, liu2025culturevlm} use internet-sourced images, while others like MaRVL \citep{liu2021visually} and CultureVQA \citep{nayak2024benchmarking} use crowdsourcing, or a combination of both, as seen in CVQA \citep{romero2025cvqa}. The selection and number of categories also remains inconsistent: CVQA adopts ten categories, CultureVQA uses only five, and CVQA’s borrows its (slightly modified) taxonomy from OK-VQA \citep{marino2019ok} which was originally designed for generic visual question answering, not cultural reasoning. We give an overview of concepts used across literature in ~\autoref{fig:combined-category-plots} and in more detail in \autoref{tab:survey-2} in Appendix.
While these choices may not be incorrect --- since the categories do reflect some culturally meaningful domains --- they reveal a lack of theoretical and historical grounding and consistency across the literature.
 \subsection{Model evaluations}

While dataset curation defines \textit{what} cultural concepts are available for analysis, model evaluations define \textit{how} well models interpret them. 
In VLMs, this has been studied within the framework of recognition, images as cultural proxy \citep{yadav2025beyond}, working with visual domains such as advertisements \citep{Zhao2024EnhancingBM}, paintings \citep{mohamed2024no, bin2024gallerygpt}, memes \citet{xu2024generating}  and gestures \citep{yerukola2025mind}.
The models used, across culturally motivated studies, is also severely fragmented, as we found in our analysis (App.~\autoref{fig:venn-plot} with more detail in ~\autoref{tab:survey_works}).
Evaluation metrics such as BLEU, ROUGE, METEOR, and CIDEr are commonly used in image-to-text tasks but often fail to capture cultural nuances due to their focus on lexical overlap, leading to mismatches with human judgments \citep{karamolegkou2024vision}. To address these limitations, researchers have proposed alternative evaluation strategies. \citet{nayak-etal-2024-benchmarking}, for instance, use LAVE—an LLM-based metric that better aligns with human reasoning \citep{10.1609/aaai.v38i5.28212}. In text-to-image evaluation, culturally grounded human protocols now assess cultural relevance, faithfulness, and realism, while metrics like the quality-weighted Vendi Score (qVS) \citep{senthilkumar2024beyond} and the Component Inclusion Score (CIS) \citep{said2025deconstructing} quantify cultural diversity and compositional accuracy, respectively.


\paragraph{Our Survey:} To find the gaps in current liteature, we examine 35 recent papers which study image(s) and culture in VLMs.
These studies curate culturally relevant datasets, evaluate multimodal models for cultural competence, and/or use images as a proxy for culture somewhere in their pipeline \footnote{The list of papers can be found in the Appendix~\ref{app:sec:litsurvey}.} This is not meant to be an exhaustive list, but rather illustrative of where the field is generally heading. we observed, a significant number of papers do not elaborate visual categories for images, evaluation frameworks, or their methodologies grounded in visual cultural studies. To address this, next we look at how visual cultural studies have examined images and how can we take inspirations to ground works in VLMs.

\section{Visuals in cultural studies}
\label{sec:visual_studies}


Visual cultural studies asks us to take seriously several aspects of cultural forms including their semiotics, materiality, gesture, and context. We draw on and expand on the digital and computational humanities research that has begun to model such an approach. Early work emerged in art history and film studies in the 1970s and 1980s focused on formal readings of art formalism and film language forcomputational analysis  \citep{lindsay1966, salt1974statistical, tsivian2009cinemetrics} Lev Manovich later argued that cultural analytics offered a situated way to study new media, specifically digital images among a quickly expanding internet \citep{manovich2020cultural}. More recently, Arnold and Tilton's theory and method of ``distant viewing'' draws on visual culture, semiotics, and media studies to theorize the cultural and social specificity of computer vision \citep{arnold2023distant}. We draw on this work to amplify the need to incorporate insights from visual cultural studies.

\subsection{Semiotically layered}
\label{sec:semiotics}
Language has direct meanings and meanings layered in how things are said or the use of certain phrases which have deeper cultural meanings. 
For example, the phrase ``apple pie'' in American English transcends its literal meaning, symbolising traditional American values and the idealised family life. Language also has stylistic variations, and misunderstandings can arise from conflicts about deeper cultural meanings of the language. Linguistics has studied the significant of these stylistic variations \citep{gao2005japanese, larina2015culture, vinay1995comparative} and cultural sensitivity it brings along \citep{pedersen2025evaluating, liu2024multilingual}.

\par

This is equally significant in images across cultures, as images can have non-literal interpretations with their meaning understood locally. In visual culture studies, this has been seen in the interpretation of visual signs and symbols from the viewpoint of cultural contexts.\citet{chandler2022semiotics}  provide a framework for decoding the layers of meaning embedded in visual media and revealing underlying cultural codes and ideologies. This semiotic tradition in cultural studies was first established by \cite{barthes1977image} and developed by 
\cite{hall1997representation}. It provides a crucial framework for understanding how images convey cultural meaning. Barthes distinguished between denotative meaning (what is literally shown) and connotative meaning (cultural associations and symbolic interpretations). For example, a pasta ad’s visual cues (tomatoes, net bag, Italian colours) signify ``Italianicity'', a cultural association that viewers can decode using shared cultural knowledge. 

Visual semiotics draws on three complementary frameworks to interpret images: a) Saussurean signifier-signified method \citep{chandler2002basics}  b) Peirce's triadic theory, and c) Panofsky's bridge between the two (\cite{panofsky1939iconology}. The latter offers a framework for analysing cultural images than the traditional Saussurean signifier-signified dyad. 
 While Saussure's model focuses on the relationship between form and meaning, i.e.~signifier and signified, Peirce examines how signs function in relation to their objects and interpretants.   Peirce's framework offers three phenomenological categories: firstness (immediate qualities), secondness (actual existence), and thirdness (interpretation). He then describes how signs relate to their objects: icons (based on resemblance), indices (based on causal connection), and symbols (based on convention). The power of this framework for cultural analysis becomes evident at the level of thirdness, where symbolic interpretation of images (or elements in images) can vary dramatically across cultures. For instance, a photograph (a type of sign) of a person making a circle with thumb and forefinger operates as a quality (firstness) by resembling the gesture, as brute fact (secondness) by documenting an actual hand position, and abstractly (thirdness) through its meaning shifts from ``OK/perfect'' in American culture to a deeply offensive gesture in Turkey or Brazil --- demonstrating how cultural knowledge fundamentally shapes interpretation at the level of thirdness \citep{long2021cultural}. 

One way to explain why identical gestures can carry vastly different meanings across cultures is to turn to Saussure. He argues for the arbitrariness of the sign, showing that the relationship between signifier and signified is culturally constructed rather than natural. His theory of value and difference reveals that meaning emerges from contrast within a system, and his synchronic analysis, or attention to change over time, emphasizes the role of historical context in shaping sign systems. Erwin Pansofsky systemized these theories within visual culture. Panofsky’s three-level iconological method complements semiotic analysis by moving from pre-iconographical description of basic visual elements, to iconographical recognition of culturally familiar symbols, and finally to iconological interpretation, which reveals deeper cultural worldviews and historical contexts. This structured approach aligns with Peirce’s concept of thirdness, offering a clear methodology for cultural analysis. Panofsky thus bridges semiotic theory and visual culture through a systematic framework for interpreting meaning in images.

\textbf{High-Context and Low-Context Cultural Messages} To understand the difference in understanding and interpretation of images, we return to Hall.  Hall stated ``high context (HC) communication is one in which most of the information is already in the person, while very little is in the coded, explicitly transmitted part of the message. A low context (LC) communication is just the opposite; i.e.,~the mass of the information is vested in the explicit code”. In high-context cultural messages, people tend to respond to make meaning constructed through straightforward and unambiguous language, often in the form of text. The more overt approach  contrasts with low-context messages, which rely on implicit and shared language alongside non-verbal cues. 

These distinctions have important implications for visual interpretation and we will see in ~\autoref{frameworks}.

\subsection{Material culture}
\label{sec:material-embodied}

Attention to material culture helps us analyze non-verbal cues. For example, the materiality of cultural objects, as outlined by \citet{tilley2005handbook}, shapes how images serve as cultural markers. These studies have produced detailed taxonomies and classification systems that could be directly adapted for VLM use. \citet{tilley2005handbook} talks about the relationship of things with human culture as \textit{``artefacts relate to the identities of individuals and groups and cultural systems of value''}. 
It provides a set of taxonomies for analyzing cultural objects in images: a)~Object types: domestic, personal, ritual, political, artistic, communal objects b)~Visual properties:~form, color, texture, decoration c)~Spatial relations:~positioning, proximity and orientation; c) Use contexts:~everyday, ceremonial d)~Symbolic elements:~religious symbols, cultural motifs.
It also gives importance to social associations --- gender-specific, age-related, status-marked, and ethnic markers; and production markers --- handmade/industrial, local/imported, and traditional/modern distinctions as they are different across cultures.
It also provides questions to understand this ``cultural inventory" in household images, which can be used as a guide for the codebook to annotate cultural images.

\subsection{Gesturally encoded}

Gestures offer another sign to study within images as conveying cultural meaning. They have been studied as a subpart of semiotics in culture, and are mentioned time and again in culture studies. There are many types of gestures that could be encoded into our models and evaluated.   
\label{sec:gestures}



For example, kinesics (the science of gestures in speech) was developed through the works of \citep{birdwhistell1970some} and \citep{ekman1969repertoire} and has become essential for cross-cultural communication studies \citep{muratova2021non}. The latter classified human interactions based on their fundamental communication
functions: Emblems (gestures, e.g., thumbs up), illustrators (bodily movements illustrating the verbal message, e.g.,~pointing), regulators (conversation flow, e.g.,~nodding), and adaptors (self-touching behaviors, e.g.,~scratching hair). Out of these, emblems are highly prone to cross-cultural interpretation \citep{matsumoto2013cultural, kendon2004gesture}, though previous works have shown differences in other non-verbal cues in cultures too \citep{argyle1994gaze, matsumoto2008culture, batty2024visual}. \citet{brunner2021multimodal} also provide a taxonomy of nonverbal features such as
gesture, facial expressions, and physical stance which could be used as a guiding factor to incorporate in multimodal data annotations. However, as recent efforts in evaluating models' ability to understand these non-verbal cues are fairly limited  \citep{lin2024meaning,yerukola2025mind}; the extent to which these models handle these cultural nuances largely remains unexplored.

\subsection{Temporal dimension}

Temporal analysis of visual materials has long been a cornerstone in visual culture studies. Researchers examine how images reflect and influence cultural narratives over time, identifying ways of seeing and tropes that shape ideas, values, and beliefs \citep{berger1973ways, sontag1977photography}. Images function not merely as passive visual objects, but as active and historical symbolic mediators that allow individuals to act upon the world, engage with others, and shape self-perception. Appadurai \citep{appadurai1988social} in his seminal work ``The Social Life of Things'', emphasizes that understanding the cultural and relational meanings of material artifacts—images included—requires attending to their social and historical trajectories, circulation, and entanglement in person-object relations. Most recently, \citet{awad2020social} studies these transformations of images in everyday life contexts to explore what kind of dynamics and influence they could have on public discourse. Attention to how images circulate over time requires us not to flatten their meaning and to take into acount how the shape ideas and believes through signs over time.




\input{sec4}
\section{Conclusion}

This paper highlights the current disconnect between technical approaches to cross-cultural multimodal learning and the rich, interdisciplinary traditions of visual cultural studies. We argue that efforts to evaluate cultural competence in VLMs remain fragmented and risk reductive representations of culture unless grounded in more holistic, nuanced frameworks.
Drawing on insights from cultural studies, semiotics, and visual studies, we provide an overview through which culture can be understood and operationalised in VLMs. 
We also point to a largely overlooked temporal dimension that warrants future attention.
Rather than proposing yet another benchmark, our goal is to revisit and re-contextualize foundational theories of culture to inform the development of more inclusive, interpretable, and culturally responsive multimodal systems. 

\section*{Limitations}

In this work, we have focused on three disciplines --- cultural studies, semiotics, and visual studies --- as representatives of a larger research area we have called visual cultural studies. This is just one path for cultural evaluations, and collaborating with the plethora of fields that animate the humanities and social sciences offers exciting paths forward. 
Our literature review used specific criteria related to visual cultural studies ---  papers which proposed dataset curation for cultural context, model evaluation to find cultural bias, probing VLMs to understand cultural knowledge gaps --- to identify relevant papers, and like with all structured literature reviews, there could be other works missing from our review.

Throughout this work, an underlying assumption has been that it is beneficial to improve the cultural competencies of VLMs, both for users (to minimize errors and harms) and for researchers (to better study cultural research questions).
However, it is important to note that this is not always (or not necessarily) true.
Sometimes, to improve a model requires training or finetuning on more data from specific cultures. Collaborative and participatory processes are one way to build models that engage with the very communities who are represented in VLMs.  Cultural theory offers one approach to more ethical, generous, and culturally-informed vision-language models.

\section*{Acknowledgments}
This work was supported in part by the Pioneer Centre for AI, DNRF grant number P1. 

Antonia Karamolegkou was supported by the Novo Nordisk Foundation (grant NNF20SA0066568) and the Onassis Foundation - Scholarship
ID: F ZP 017-2/2022-2023.



\bibliography{custom}



\newpage

\appendix

\section{Technical Appendices and Supplementary Material}

\subsection{Frequency of Model Use In Cultural Evaluation}
\label{sec:model-frequency-use}


\begin{figure}[h]
    \centering
    \includegraphics[width=1\linewidth]{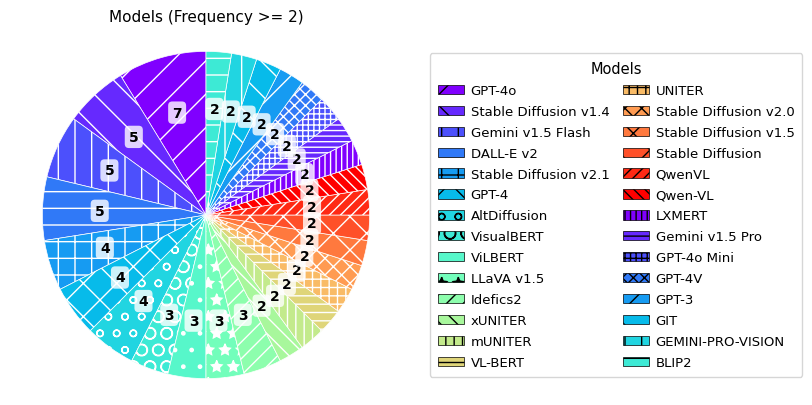}
    \caption{Plot showing the frequency of use of models for evaluation of cultural knowledge. All the models that have a frequency of less than 2 are not represented, there are 126 such models. The plot demonstrates the variability of models used across papers.}
    \label{fig:venn-plot}

\end{figure}

\newpage

\subsection{Proposed questions for studying advertisement complexity}
\label{sec:ad-questions}

\begin{table*}[h]
    \centering
    \footnotesize
    \begin{tabular}{p{4cm}p{4cm}p{4cm}}
        \toprule
        \textbf{Original Score Item \newline \citet{hornikx2017influence}} & \textbf{Proposed Direction for Measurable Questions in Images} & \textbf{What it Tests About the Model} \\
        \midrule
        Speakers should not expect that listeners will figure out what they really mean unless the intended message is stated precisely. & Does the image clearly describe what the advertisement is about, with no ambiguity? & Tests how explicitly the model communicates action and meaning. \\
        \midrule
        It is more important to state a message efficiently than with great detail. & Does the image prioritise brevity over completeness? & Measures tradeoff between efficiency and contextual richness. \\
        \midrule
        Even if not stated exactly, a speaker’s intent will rarely be misunderstood. & Step 1: Is the advertisement's intent explicitly stated? |
        Step 2: If not, is the intent still understandable? & Tests robustness of implied intent. \\
        \midrule
        Intentions not explicitly stated can often be inferred from the context. & Step 1: Is there advertisements' intentions implied rather than stated? | Step 2: If yes, do visual or cultural cues help in understanding the intent? & Evaluates model’s ability to embed inferable meaning through context. \\
        \midrule
        A speaker can assume that listeners will know what they really mean. & Step 1: Do you understand the meaning of the advertisement? | Step 2: If yes, what elements of the image helped in this knowledge, including associations to our culture? & Measures reliance on shared knowledge; detects cultural associations or omissions. \\
        \midrule
        People understand many things that are left unsaid. & Step 1: Are key cultural or social meanings implied rather than named? | Step 2: If yes, can you state these things and interpret them? & Tests model’s capacity to rely on implicit symbolic communication. \\
        \midrule
        Fewer words can often lead to better understanding. & Does the model use minimal language while still preserving clarity? & Tests if the model maintains accuracy through brevity. \\
        \midrule
        You can often convey more information with fewer words. & Step 1: Are there terms, gestures, things, or phrases that carry deeper or layered meaning? | Step 2: If yes, what meanings are conveyed implicitly through those elements? & Measures semantic richness and cultural symbolism. \\
        \midrule
        Some ideas are better understood when left unsaid. & Step 1: Is there a cultural, emotional, or symbolic message that is implied but not stated? | Step 2: Would directly stating that message weaken the impact of the advertisement? & Evaluates model’s use of subtlety, emotional tone, and symbolic omission. \\
        \bottomrule
    \end{tabular}
    \caption{Proposed questions to test the visual complexity of advertisements, taken from \citet{hornikx2017influence}. A high score implies association with high context culture, except for the first item}
\end{table*}

Proposed questions to test ad's visual complexity taken from \citet{hornikx2017influence}, which itself is a subset of questions from \citet{richardson2007influence}. Responses to these questions can be measured on 5 point scale as done in \citet{hornikx2017influence}.

\newpage

\subsection{Example image framing in ads in culture}
\label{sec:ad-framing}

\begin{figure}[h]
  \centering
  \includegraphics[width=0.7\linewidth]{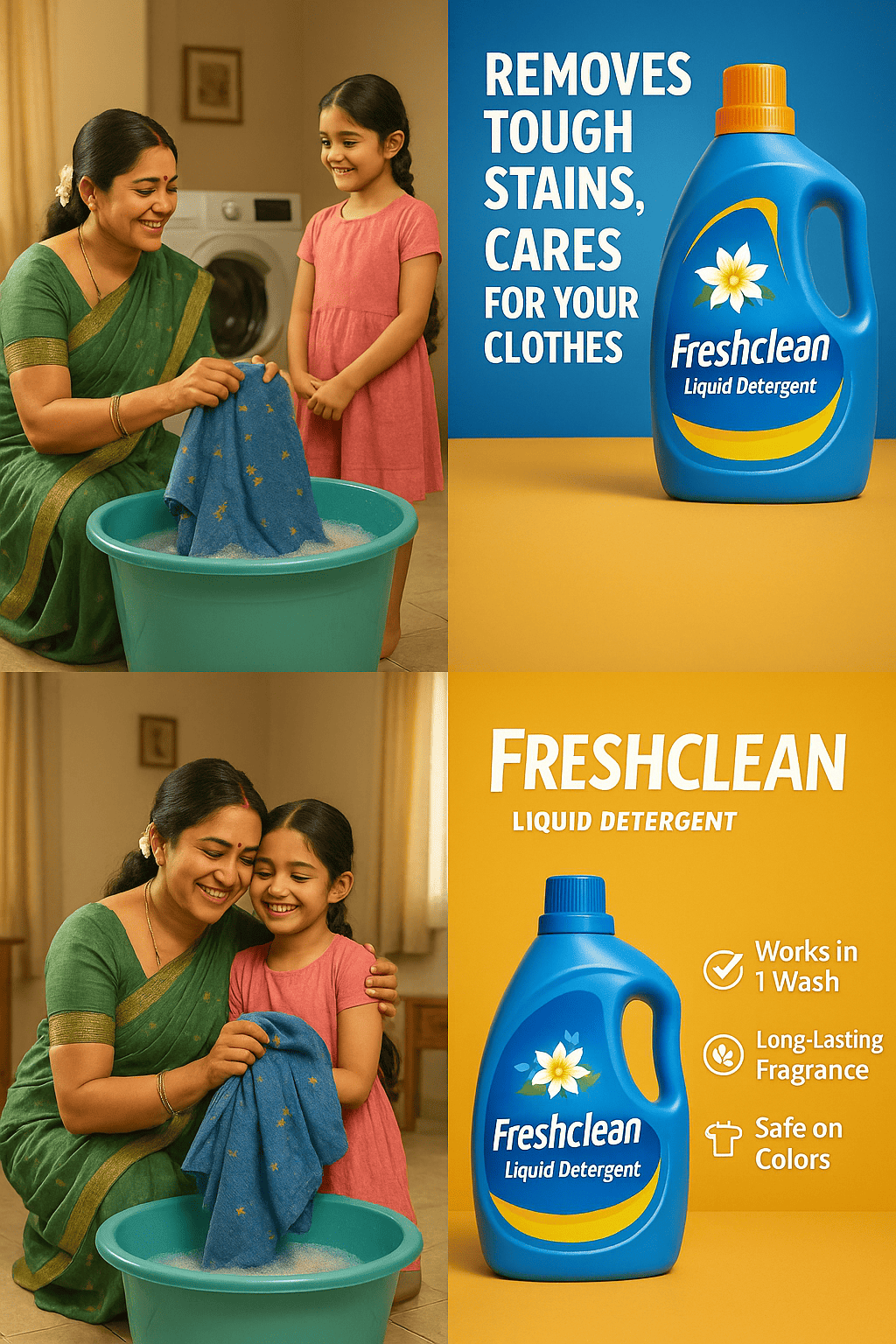} 
  \caption{An AI-generated advertisement (with prompt `draw a liquid detergent advertisement for India' showing that images in application differ in cultural context message across cultures. For example, we have a high-context country(India) in the image. As per cultural theory (~\autoref{sec:semiotics}), in high context cultures (e.g. India) the images have little in the coded, explicitly transmitted part of the message, which is contrary to what we see here in the image.} 
  \label{fig:ai-ads}
\end{figure}

\newpage 

\subsection{Mapping of all categories used in literature}
\label{sec:}

\begin{figure}[h]
  \centering
    \includegraphics[width=0.5\linewidth]{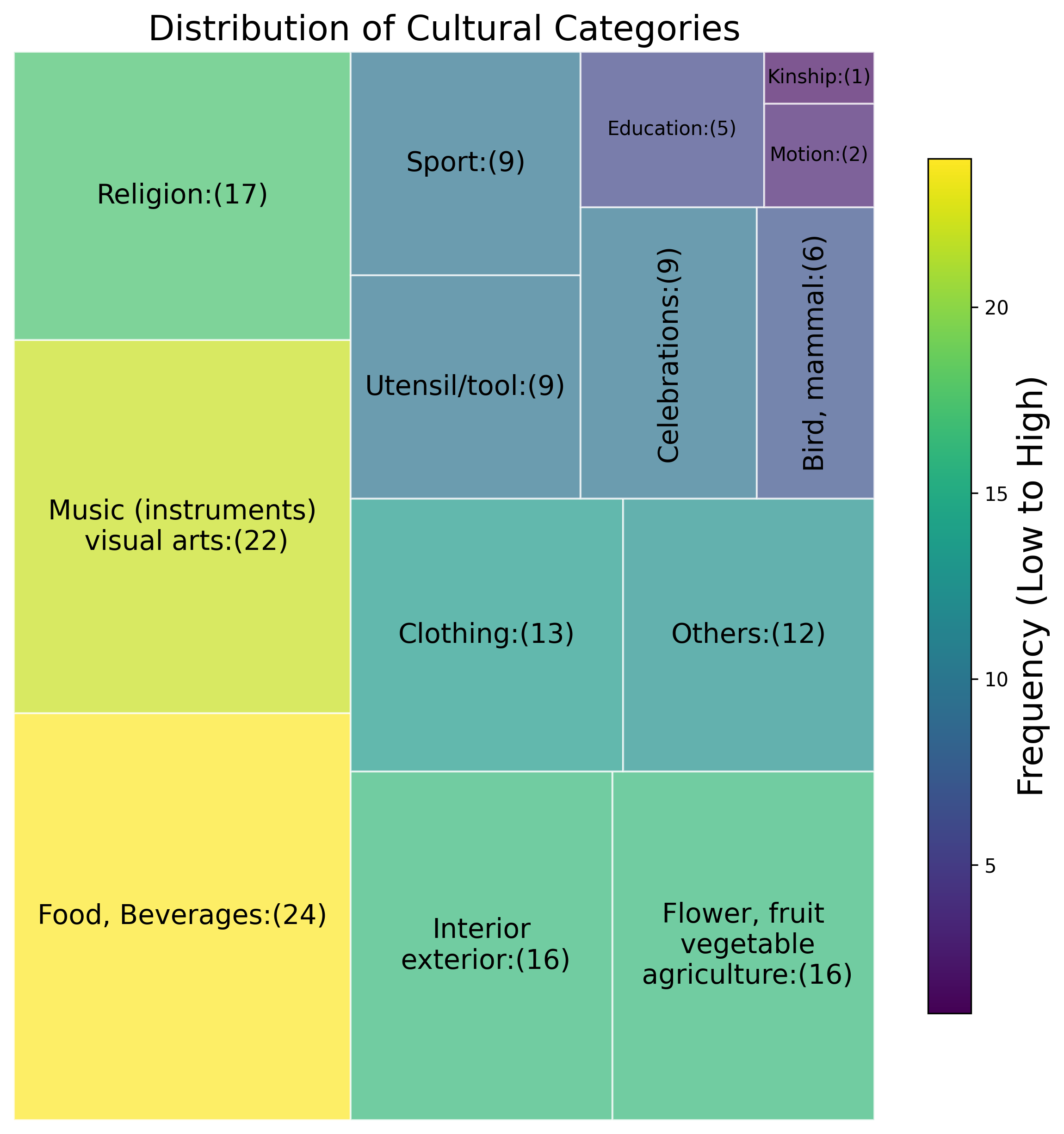}
    \caption{Overview of the distribution of concepts across the literature studied (Appendix~A.6). Each block represents a concept from IDS \citep{key2015ids} semantic concepts after we mapped all categories used in literature ($\sim 100$) to IDS categories. We observe that some concepts like food, visual arts are studies way more than others like utensils and celebrations. See Table~\ref{tab:survey-2} in the Appendix for the full list of papers and their attributes. }
  \label{fig:combined-category-plots}
\end{figure}

\clearpage
\newpage

\subsection{List of surveyed works}
\label{app:sec:litsurvey}

In \autoref{tab:survey_works} we present an overview of the works we examined, focusing on aspects such as cultural \textit{concepts}\footnote{We note that some papers that study culture in VLMs don't explicitly mention \textit{concept} as a term. They can be called broad categories, domains, etc. We see their intent of using these categories/domains and place them under the concepts column.}, theoretical grounding, and model fine-tuning. The majority of studies include material and embodied culture, with some also considering semiotically layered elements. While many papers have mentioned what broad cultural cultural concepts (details of which can be found in ~\autoref{tab:survey-2}) they used in either data curation or evaluations , there's a noticeable variation in the depth of theoretical grounding, ranging from explicit frameworks (suggested by \citet{key2015ids} and \citet{halpern1955dynamic}) to absent theoretical references (NA). The models evaluated span both fine-tuned and non-fine-tuned variants.
\input{tables/survey}

\input{tables/survey-table2}

\newpage

\end{document}

%% file: commands.tex
\definecolor{maria-color}{HTML}{7881F2}

\definecolor{srishti-color}{HTML}{AC78F2}

\definecolor{antonia-color}{HTML}{556B2F}



%% file: sec4.tex
\section{A set of grounded frameworks for cross-cultural VLMs}
\label{frameworks}

To evaluate and build culturally-aware vision-language models (VLMs), we propose a set of five frameworks rooted in visual cultural studies. Each individual framework corresponds to a theoretical insight and methodological shift: from how images are interpreted and who interprets them, to what kinds of cultural objects and meanings are encoded, and how they evolve over time. This section unpacks each component — processual grounding, material culture, symbolic encoding, contextual interpretation, and temporality — and shows how it informs core tasks in multimodal learning, such as cultural captioning, contrastive evaluation, dataset design, and text-to-image generation.

\subsection{Processual grounding: who defines culture?}

A critical challenge in designing and evaluating VLMs for cultural competence is the question of who defines and judges cultural meaning. This is closely tied to the anthropological distinction between emic and etic perspectives: cultural evaluations should ideally be judged by insiders (emic), rather than imposed by external observers (etic) \citep{pike1967language,harris1968rise}.

Participatory visual methodologies offer an answer. In photo elicitation \citep{harper1988visual}, researchers use images to elicit cultural meaning from participants themselves. Similarly, Photovoice \citep{wang1997photovoice} invites communities to produce and reflect on their own visual representations. Both methods flip the conventional direction of evaluation: the cultural meaning is not imposed post-hoc but defined from within.
Such approaches are especially relevant for generative models. Current evaluations typically prompt T2I models with cultural keywords and then assess the outputs against fixed templates. Instead, participatory pipelines could ask annotators or communities to specify prompts, select relevant outputs, or even co-create evaluation rubrics — ensuring that cultural fidelity is judged through lived knowledge.

Further, perception itself is culturally shaped. \citet{nisbett2005influence} show that East Asian viewers often attend more to context and relationships in visual scenes, whereas Western viewers focus on focal objects. Such differences affect how people annotate or describe images. \citet{ye2023computer} show that captions written by native speakers differ not just in fluency but in semantic framing. This calls for enregisterment-aware evaluation, where aesthetic and linguistic features (e.g., lighting, style, register) are analyzed as part of the cultural signal \citep{nakassis2023linguistic}.

Panofsky’s iconological hierarchy \citep{panofsky1939iconology} further enriches this framing with three levels of image understanding:
(1) \textit{Pre-iconography}: literal description of visual elements; (2) \textit{Iconography}: cultural decoding of symbols and known references; (3) \textit{Iconology}: deep, contextual interpretation tied to worldview and history. This three-level lens helps separate superficial model success (e.g., labeling objects) from deeper cultural competence.

Current tasks that can benefit from this perspective include participatory evaluation of T2I generation, especially through \textit{community-authored captioning datasets}, \textit{emic-driven prompt design}, and \textit{annotation protocols that account for perceptual framing} (e.g., holistic vs. analytic viewing). Moreover, tasks involving \textit{captioning or cultural QA}—inspired by Panofsky's pre-iconographic to iconological hierarchy—can assess the depth of cultural understanding encoded in captions or retrieval outputs. 



\subsection{Material and embodied culture: what is represented?}
Cultural meaning is not only symbolic — it is also embedded in physical forms, spatial arrangements, and production contexts. Drawing on material culture theory \citep{tilley2005handbook}, we argue for expanding the scope of visual annotations to include not just what is in the image, but further metadata (i.e. how it is made, used, and socially situated). Datasets from \citet{li2024foodieqa} and \citet{magomere2024} have begun collecting fine-grained metadata: meal types, utensils, cooking methods, taste, color, or associated occasions. Yet even these stop short of capturing the full embodied and contextual affordances of cultural objects. We build on this by proposing a taxonomy of attributes (~\autoref{tab:cultural-object-taxonomy}) including:
Object Types (e.g., ritual vs. domestic), Visual Properties (e.g., form, material), Spatial Relations (e.g. positionality), Use Contexts (e.g., everyday vs. ceremonial), Social Associations (e.g., gender, class), Production Markers (e.g., handmade vs. industrial). These provide richer ways to evaluate cultural understanding beyond simple recognition. A dish isn't just “food” — it is a marker of status, identity, or ritual meaning.

This dimension supports tasks that go beyond surface recognition, such as \textit{dataset curation for cultural QA, retrieval, or classification}, \textit{fine-grained captioning, detection, and reasoning over objects-in-context}, and \textit{commonsense QA involving usage, production, or social context}. It also informs \textit{dataset construction pipelines} that embed material and social metadata—ranging from production markers to social usage—into annotation schemas.






\begin{table}
    \centering
    \scriptsize
    \parbox{.45\linewidth}{
    \begin{tabular}{p{2.2cm}p{3.5cm}}
        \toprule
        \textbf{Category} & \textbf{Details} \\
        \midrule
        \textbf{Object Types} & Domestic, personal, ritual, political, artistic, communal \\
        \addlinespace
        \textbf{Visual Properties} & Form, color, texture, decoration \\
        \addlinespace
        \textbf{Spatial Relations} & Positioning, proximity, orientation \\
        \addlinespace
        \textbf{Use Contexts} & Everyday use, ceremonial or ritual settings \\
        \addlinespace
        \textbf{Symbolic Elements} & Religious symbols, cultural motifs \\
        \midrule
        \textbf{Social Associations} & Gender-specific, age-related, status-marked, ethnic markers \\
        \addlinespace
        \textbf{Production Markers} & Handmade vs. industrial, local vs. imported, traditional vs. modern \\
        \bottomrule
    \end{tabular}
    \vspace{0.3cm}
    \caption{Taxonomy for analyzing cultural objects in images, based on attributes and sociocultural associations. 
    }
    \label{tab:cultural-object-taxonomy}
    }
    \hspace{0.8cm}
    \parbox{.45\linewidth}{
    \begin{tabular}{p{2cm} p{3.5cm}}
        \toprule
        \textbf{Component} & \textbf{White Dove Example} \\
        \midrule
        \textbf{Saussurean Layer} & 
        Denotative: ``A white dove flying in the sky.''  
        Connotative: ``Signifies peace and hope.''\\
        \addlinespace
        \textbf{Peircean Layer} & 
        Firstness: Bird silhouette resembles actual dove.  
        Secondness: Seen at a peace rally, indicating a real event.  
        Thirdness: Universally read as a symbol of peace and reconciliation.\\
        \addlinespace
        \textbf{Benchmark Task} & 
        Contrastive captions—(a) “A white bird in flight.” vs.  
        (b) ``A dove symbolising peace at the peace rally.''  
        Ask the model to pick which conveys the cultural meaning.\\
        \bottomrule
    \end{tabular}
    \vspace{0.3cm}
    \caption{Proposed framework to study \textit{symbolism} involving both annotation schema and evaluation task.}
    \label{tab:semiotic-example-schema}
    }
\end{table}

\subsection{Symbolic and semiotic encoding: how is meaning layered?}
Culture operates heavily through symbolism, and images often encode multiple layers of meaning: what is shown (denotation), what it evokes (connotation), and how it is interpreted within cultural systems. This aligns with the semiotic frameworks of \textit{the signifier} (the form of a word or image) and \textit{the signified} (the concept it represents) \citep{saussure1916cours} and the triadic model: \textit{the representamen} (the form), \textit{the object} (what it refers to), and \textit{the interpretant} (the meaning derived) \citep{peirce1868categories}; further elaborated by cultural contexts in distinguishing between \textit{denotation} (literal meaning) and \textit{connotation} (cultural meaning)\citep{barthes1977image}.
For instance, a white dove may denote “a bird in flight” but connote “peace” or “hope,” depending on context. Peirce’s triadic model helps evaluate models at three levels: (1) Firstness: appearance (is the visual form recognizable?) (2) Secondness: referential grounding (is it used in a real context?) (3) Thirdness: symbolic interpretation (does it evoke a shared cultural idea?)

Most current models succeed at the first, are inconsistent at the second, and rarely reach the third. To address this, we propose a multi-layer annotation schema (see Table~\ref{tab:semiotic-example-schema}) and corresponding benchmark tasks: e.g.,~choosing between contrastive captions that test literal vs.~symbolic interpretations.
Visual metaphors and gestures are particularly rich in this regard. A hand gesture might be read as ``OK'' in one culture and as an insult in another --- a classic example of Peircean thirdness. Yet such signals remain largely untested in VLM evaluation.

 The theory of `high-context' and `low-context` prominently affects the advertisement space. Ads are intentionally made to appeal to different cultures.  With the rise of AI generated content, one of the sought-after uses of text-to-image (T2I) models is in creating visually appealing and persuasive ads \citep{aghazadeh2024cap}. From the perspective of cross-cultural awareness of these models, there lack a thorough study of whether today's models fully incorporate deep culturally-specific context when generating these images (App. ~\autoref{fig:ai-ads}). We also propose a framework to study this complexity using context score \citep{hornikx2017influence} (details in ~\autoref{sec:ad-questions}). We need to redefine how we evaluate models with attention to pragmatics. We also need to be careful; this is not an argument about high and low culture. As cultural studies scholar Raymond Williams reminds us, it is the everyday lived experience that shapes culture~\citep{williams1983culture}. 

Tasks that benefit from this perspective include \textit{contrastive caption evaluation} (e.g.,~distinguishing literal vs.~symbolic interpretations), \textit{cross-cultural symbol},  \textit{ad generation using T2I models} and \textit{visual metaphor detection and interpretation}. This perspective also aids \textit{bias detection frameworks} that probe where models fail to grasp symbolic nuance or misread culturally loaded signs.





\subsection{Contextual interpretation: who understands and frames meaning?}
Cultural meaning is not fixed — it is shaped by the viewer’s background, assumptions, and interpretive frame. Drawing on Hall’s encoding/decoding model \citep{hall2007encoding}, we highlight the need for evaluations that account for diverse perceptual framing. We extend this to propose a three-layer evaluation approach (Table~\ref{tab:evaluation-framework-sep}) based on: (1) \textit{Who evaluates} (emic vs. etic); (2) \textit{Who defines the cultural frame} (e.g., user-guided prompting); (3) \textit{How meaning is read} (framing effects, stereotype activation).
\begin{wraptable}{L}{0.48\textwidth}
    \centering
    \scriptsize
    \renewcommand{\arraystretch}{1.0}
    \begin{tabular}{p{0.22\linewidth} p{0.25\linewidth} p{0.3\linewidth}} 
    \toprule
    \textbf{Dimension} & \textbf{Core Question} & \textbf{Focus of Evaluation} \\
    \midrule
    \textbf{Emic vs. Etic} &
    Who judges the cultural fidelity of the generated image? &
    \textit{Who evaluates the output?} \\
    \addlinespace
    \textbf{Inward Participation} &
    Who defines what cultural elements should be depicted? &
    \textit{Who defines the input and cultural frame?} \\
    \addlinespace
    \textbf{Perceptual Framing} &
    How might different viewers interpret the same image? &
    \textit{How meaning is interpreted?} \\
    \bottomrule
    \end{tabular}
    \caption{Three-layer framework to study \textit{contextual interpretation} for culturally grounded evaluation of T2I models.}
    \label{tab:evaluation-framework-sep} 
    \vspace*{-0.5cm}
\end{wraptable}

Most evaluations currently center on the model or the annotator. Instead, we argue for an insider-led prompting and evaluation (an insider would be a person who identifies with the examined culture). For example, in T2I, one could ask all the annotators to generate images of their own culture and then reflect on model accuracy, shifting agency to communities. This lens also requires accounting for high-context vs. low-context cultural communication styles \citep{hall1997representation}, and whether models reflect pragmatic, implicit knowledge beyond literal interpretation. 

Here, tasks include \textit{insider-led T2I evaluation protocols}, \textit{cross-cultural user studies for multimodal perception}, and \textit{comparison of model vs. human performance}. It also supports \textit{pragmatic reasoning tasks} that depend on sociocultural inference and context-aware understanding.






\subsection{Temporality: when is culture situated?}
Culture is dynamic ---  what something means changes over time. Yet, most VLM evaluations treat culture as static, reducing it to fixed objects or timeless categories. Cultural studies, by contrast, emphasize the historicity of meaning \citep{appadurai1988social, awad2020social}.
Visual media from different eras encode shifting aesthetics, values, and norms. For instance, newspaper advertisements from the 1950s differ significantly in racial and gender representation compared to today. Such diachronic data can be used to evaluate whether VLMs are sensitive to cultural change or trapped in frozen stereotypes.

Datasets such as the Newspaper Navigator\footnote{https://huggingface.co/datasets/biglam/newspaper-navigator} or art collections\footnote{https://www.workwithdata.com/datasets/artworks?f=1\&fcol0=art\_medium\&fop0=\%3D\&fval0=Painting} spanning decades, offer a promising testbed. Future benchmarks could track how models interpret symbols across time and flag inconsistencies or anachronisms. Temporal awareness enables tasks such as \textit{diachronic evaluation of cultural representation and symbols}, \textit{image generation across time periods}, and \textit{longitudinal cultural QA} using archives or historical art corpora. It also informs \textit{bias audits over time}, which evaluate whether VLMs reflect shifts in cultural symbols, norms, and aesthetics.

%% file: tables/survey.tex
\begin{longtable}{p{0.20\textwidth} p{0.15\textwidth} p{0.10\textwidth} p{0.10\textwidth} p{0.15\textwidth} p{0.25\textwidth}}

\label{tab:survey_works}\\

\caption{Overview of papers indicating culture category, presence of clearly defined cultural concepts, theoretical grounding, and the use of fine-tuned or non-fine-tuned models.} \\
\toprule
Paper & Category & Clear-Cultural-Concept-Mentioned & Theoretical-Grounding & Finetuned Models& Not Finetuned Models \\* \midrule
\endfirsthead
\multicolumn{6}{c}%
{{\bfseries Table \thetable\ continued from previous page}} \\
\toprule
Paper & Category & Broad-Cultural-Concepts-Mentioned
& Theoretical-Grounding & Finetuned Models & Not Finetuned Models\\* \midrule
\endhead
\bottomrule
\endfoot
\endlastfoot
Visually Grounded Reasoning across Languages and Cultures \citep{liu2021visually} & material culture & Y & IDS \citep{key2015ids} & UNITER, mUNITER, xUNITER, VL-BERT, VisualBERT, ViLBERT, LXMERT & None \\
\midrule\\
Evaluating Visual and Cultural Interpretation: The K-Viscuit Benchmark with Human-VLM Collaboration \citep{baek2024evaluating} & material culture & Y & IDS \citep{key2015ids} & None & InstructBLIP, mPLUG-Owl2, LLaVA-1.6, InternLM-XC2, Molmo, Idefics2, Llama-3.2, Claude-3, GPT-4, Gemini-1.5. \\
\midrule\\
CVLUE: A New Benchmark Dataset for Chinese Vision-Language Understanding Evaluation\citep{wang2025cvlue} & material culture & Y & NA & CCLM, X2VLM & QwenVL, QwenVL-Chat, mPLUG-Owl2 \\
\midrule\\
FoodieQA: A Multimodal Dataset for Fine-Grained Understanding of Chinese Food Culture \citep{li2024foodieqa} & material culture/ semiotically layered & Y & Chinese Cuisines \citep{Zhang2020} & None & GPT-4, GPT-4V, GPT-4o, Phi-3-vision, Phi-3-medium, Idefics2, Mantis, Qwen-VL, Yi-VL, Llama3, Mistral, Qwen2 \\
\midrule\\
CVQA: Culturally-diverse Multilingual Visual Question Answering Benchmark \citep{romero2025cvqa} & material culture & Y & NA & None & CLIP, M-CLIP, LLaVA-1.5, mBLIP-mT0, mBLIP-BLOOMZ, InstructBLIP, GPT-4o, Gemini-1.5-Flash \\
\midrule\\
On the Cultural Gap in Text-to-Image Generation \citep{liu2024cultural}  & material culture & N & NA & Stable Diffusion v1-4 & Stable Diffusion v1-4 \\
\midrule\\
Semantic and Expressive Variation in Image Captions Across Languages \citep{ye2023computer}  & semiotically layered & N & \citep{Nisbett2003,Cenek2015} & GIT & LLaVA, Google Vertex API imagetext-001 \\
\midrule\\
ViTextVQA: A Large-Scale Visual Question Answering Dataset for Evaluating Vietnamese Text Comprehension in Images
\citep{van2024vitextvqa}  & material culture & N & NA & M4C, BLIP-2, LaTr, SaL, PreSTU, ViTextBLIP-2 & GPT-4o, Gemini-1.5-Flash, QwenVL \\
\midrule\\
HaVQA: A Dataset for Visual Question Answering and Multimodal Rese \citep{parida2023havqa}  & material culture & N & NA & BEiT-large, ViT-base, ViT-large, DeiT-base, M2M-100 & None \\
\midrule\\
Navigating Text-to-Image Generative Bias across Indic Languages\citep{mittal2024navigating} & material culture & N & NA & None & Stable Diffusion, AltDiffusion, Midjourney, Dalle3 \\
\midrule\\
The World Wide Recipe: A community-centred framework for fine-grained data collection and regional bias operationalization \citep{magomere2024} & material culture & Y & NA & None & DALL-E 2, DALL-E 3, Stable Diffusion v2.1, LLaVA-1.6 \\
\midrule\\
SEA-VQA: Southeast Asian Cultural Context Dataset For Visual Question Answering \citep{urailertprasert2024sea} & material culture & N & UNESCO & None & GPT-4, GEMINI-PRO-VISION \\
\midrule\\
Extracting Cultural Commonsense Knowledge at Scale CCSK \citep{nguyen2023extracting} & material culture & Y & NA & None & GPT-3 \\
\midrule\\
Benchmarking Vision Language Models for Cultural Understanding \citep{nayak2024benchmarking}) & material culture & Y & NA & None & GPT-4o, CLAUDE 3.5, GEMINI-PRO-VISION, BLIP2, INSTRUCTBLIP, MBLIP, PALLIGEMMA, LLaVA1.5, LLaVA-NEXT, IDEFICS2, Intern-VL 1.5 \\
\midrule\\
Broaden the Vision: Geo-Diverse Visual Commonsense Reasoning \citep{yin2021broaden}  &  material culture & N & NA & VisualBERT, ViLBERT & None \\
\midrule\\
Towards Equitable Representation in Text-to-Image Synthesis Models with the Cross-Cultural Understanding Benchmark (CCUB) Dataset \citep{liu2023towards} &  material culture & Y & \citep{halpern1955dynamic, key2015ids} & Stable Diffusion (fine-tuned on CCUB), GPT-3 (fine-tuned on CCUB captions) & Stable Diffusion, Stable Diffusion 2.0 \\
\midrule\\
AltDiffusion: A Multilingual Text-to-Image Diffusion Model \citep{ye2024altdiffusion} &  material culture & Y & NA &  AltDiffusion & Stable Diffusion v2.1, Taiyi-Bilingual, Japanese Stable Diffusion \\
\midrule\\
Partiality and Misconception: Investigating Cultural Representativeness in Text-to-Image Models \citep{10.1145/3613904.3642877} & material culture & Y & NA & DALL-E v2, Stable Diffusion v1.5, Stable Diffusion v2.1 \\
\midrule\\
Multilingual Conceptual Coverage in Text-to-Image Models \cite{saxon-wang-2023-multilingual} & material culture & N & NA & None & DALL-E Mini, DALL-E Mega, DALL-E 2, Stable Diffusion 1.1-1.4, Stable Diffusion 2, CogView2, AltDiffusion \\
\midrule\\
Navigating Cultural Chasms: Exploring and Unlocking the Cultural POV of Text-To-Image Models \citep{ventura_chasms}& material culture/ semiotically layered & N & \citep{hofstede1983national,wvs2022wave7,rokeach2006values} & None & StableDiffusion 2.1v, StableDiffusion 1.4v, AltDiffusion, DeepFloyd, DALL-E, LLaMA 2 + SD 1.4 UNet \\
\midrule\\
Beyond aesthetics: Cultural competence in text-to-image models - CUBE \citep{senthilkumar2024beyond} & material culture & Y & NA & None & Imagen 2, Stable Diffusion XL (SDXL), Playground, Realistic Vision \\
\midrule\\
Globally Inclusive Multimodal Multitask Cultural Knowledge Benchmarking \citep{schneider2025gimmickgloballyinclusive} & material culture & Y & NA & None & Claude 3.5, Gemini 1.5, GPT-4o, InternVL2.5, Qwen2 VL, LLaMA 3.2, MiniCPM V, Centurio, Qwen-VL, Idefics2, Qwen2.5, InternLM2.5, Aya-Expanse, Phi 3.5, MiniCPM LLM, Qwen2. \\
\midrule\\
WorldCuisines:A Massive-Scale Benchmark for Multilingual and Multicultural Visual Question Answering on Global Cuisines \citep{winata-etal-2025-worldcuisines} & material culture & Y & NA & None & LLaVA-1.6, Qwen2-VL-Instruct, Llama-3.2-Instruct, Molmo-E, Molmo-D, Molmo-O, Pangea, Aria, Phi-3.5-Vision, Pixtral, NVLM-D, GPT-4o, GPT-4o-Mini, Gemini-1.5-Flash. \\
\midrule\\
An image speaks a thousand words, but can everyone listen?On image transcreation for cultural relevance \citep{khanuja-etal-2024-image} & material culture & Y & \citep{key2015ids} & None & InstructPix2Pix, InstructBLIP-FlanT5-XXL, PlugnPlay, GPT-3.5, GPT-4, GPT-4o, DALLE-3 \\
\midrule\\
GeoDE: a Geographically Diverse Evaluation Dataset for Object Recognition \citep{NEURIPS2023_d08b6801} & material culture & N & NA & ResNet50 & CLIP, ResNet50 \\
\midrule\\
CIC: A framework for Culturally-aware Image Captioning \citep{youngsik_cic} & material culture & Y & \citep{halpern1955dynamic} & None & GIT, CoCa, BLIP2, ChatGPT \\
\midrule\\
ViSAGe: A Global-Scale Analysis of Visual Stereotypes in Text-to-Image Generation \citep{jha-etal-2023-seegull} & material culture/  semiotically layered & N & NA & None & Stable Diffusion v1.4 \\
\midrule\\
DIG In: Evaluating Disparities in Image Generations with Indicators for Geographic Diversity \citep{hall2023dig} & material culture/  semiotically layered & N & NA & None & LDM 1.5 (Open), LDM 2.1 (Open), LDM 2.1 (Closed), GLIDE, DM w/ CLIP Latents \\
\midrule\\
Exploiting Cultural Biases via Homoglyphs in Text-to-Image Synthesis \citep{struppek_exploiting} & material culture & Y & NA & Stable Diffusion v1.5 & DALL-E 2, Stable Diffusion v1.5 \\
\midrule\\
Inspecting the Geographical Representativeness of Images from Text-to-Image Models \citep{10376690} & material culture & Y & NA & None & DALL·E 2, Stable Diffusion \\
\midrule\\
Easily Accessible Text-to-Image Generation Amplifies Demographic Stereotypes at Large Scale \citep{bianchi_easily} & material culture/  semiotically layered & N & NA & None & Stable Diffusion v1.4, DALL·E 2 \\
\midrule\\
Cultural Concept Adaptation on Multimodal Reasoning \citep{li-zhang-2023-cultural} & material culture & Y & NA (filtering ConceptNET and WordNET) & None & mUNITER, xUNITER, UC2, M3P \\
\midrule\\
CROPE: Evaluating In-Context Adaptation of Vision and Language Models to Culture-Specific Concepts \citep{nikandrou-etal-2025-crope} & material culture & Y &IDS(MaRVL), MCC & None & LLaVA-1.5, MOLMO, LLaVA-NeXT, Phi-3-Vision-128K, Llama-3.2-Vision, Paligemma, XGen-MM-Interleaved, Idefics2, Mantis-Idefics2, VILA, InternLM-XComposer-2.5, LLaVA-OneVision, Qwen2-VL, mPLUG-Owl3, Gemini-1.5 Flash, Gemini-1.5 Pro, GPT-4o. \\
\midrule\\
Exploring Visual Culture Awareness in GPT-4V: A Comprehensive Probing \citep{cao2024exploring} & material culture & Y & \citep{key2015ids} & None & GPT-4V \\
\midrule\\
RAVENEA: A Benchmark for Multimodal Retrieval-Augmented Visual Culture Understanding \citep{li2025ravenea} & material culture & Y & NA & SigLIP2, CLIP, VisualBERT, VL-T5, LLaVA-OneVision & LLaVA-OneVision, DeepSeek-VL2, Qwen2.5-VL, InternVL3, Gemma3, Phi4-Multimodal, Pixtral, GPT-4.1 \\
\bottomrule

\end{longtable}
\newpage

%% file: tables/survey-table2.tex
\subsection{Concept used across the literature of VLMs which study culture.}

 In \autoref{tab:survey-2}, we highlight the concepts\footnote{We note that some papers that study culture in VLMs don't explicitly mention \textit{concept} as a term. They can be called broad categories, domains, etc. We see their intent of using these categories/domains and place them under the concepts column.} and fine-grained details of those concepts studied in the works that we surveyed. If the fine grained details were too many to fit in the table, we mention the table/section of the respective paper in which the information can be found. One of the observations is that there is little consistency in the concept chosen, as well as the fine-grained details that the community has looked into.

\begin{longtable}{p{0.20\textwidth} p{0.10\textwidth} p{0.35\textwidth}p{0.30\textwidth}}

\label{tab:survey-2}\\

\caption{Overview of papers indicating whether they define clear cultural concepts, along with the specific concept categories and any fine-grained details provided.} \\
\toprule
Paper & Clear-Cultural-Concept-Mentioned & Concepts & Finegrained Details\\* \midrule
\endfirsthead
\multicolumn{4}{c}%
{{\bfseries Table \thetable\ continued from previous page}} \\
\toprule
Paper & Clear-Cultural-Concept-Mentioned & Concepts & Finegrained Details\\* \midrule
\endhead
\bottomrule
\endfoot
\endlastfoot
Visually Grounded Reasoning across Languages and Cultures \citep{liu2021visually} & Y & "animal", "food and beverages", "clothing and grooming", "the house", "agriculture and vegetation", "basics and technology", "motion", "time", "cognition", "speech and language","religion and belief" & --- \\
\midrule\\

Evaluating Visual and Cultural Interpretation: The K-Viscuit Benchmark with Human-VLM Collaboration \citep{baek2024evaluating} & Y & "food", "beverage", "game", "celebrations", "religion", "tool", "clothes", "heritage", "architecture", and "agriculture". & Traditional Clothing, Historical and Social Roles, Values and Proverbs, University Culture, Culinary Culture, Traditional Beverages, Traditional Practices and Activities, Traditional Games and Entertainment, Cultural Ceremonies, Historical Figures and Structures \\

\midrule\\

CVLUE: A New Benchmark Dataset for Chinese Vision-Language Understanding Evaluation \citep{wang2025cvlue} & Y & "Animals", "food", "drinks", "clothes", "plants", "fruits", "vegetables", "agriculture", "tools", "furniture", "sports", "celebrations", "education", "musical instruments", "art" & --- \\

\midrule\\

FoodieQA: A Multimodal Dataset for Fine-Grained Understanding of Chinese Food Culture \citep{li2024foodieqa} & Y & "food" & Food from:Sichuan, Cantonese (Guangdong), Shandong, Jiangsu, Zhejiang, Fujian, Hunan, Anhui, Northwest, Northeast, Xinjiang, Jiangxi, Mongolian, and Hakka cuisines based on the categories:dish name, alternative names, food category, main and secondary ingredients, ingredient characteristics, flavor, color, presentation style, serving temperature, dishware, cooking techniques, regional origin, cuisine type, eating habits \\

\midrule\\

CVQA: Culturally-diverse Multilingual Visual Question Answering Benchmark \citep{romero2025cvqa} & Y & "Vehicles and Transportation", "Cooking and Food", "People and Everyday Life", "Sports and Recreation", "Plants and Animals", "Objects, Materials, and Clothing", "Brands and Products", "Geography, Buildings, and Landmarks", "Tradition, Art, and History", "Public Figures and Pop Culture" & --- \\

\midrule\\

On the Cultural Gap in Text-to-Image Generation \citep{liu2024cultural} & & N & --- \\

\midrule\\

Semantic and Expressive Variation in Image Captions Across Languages \citep{ye2023computer} & N & N & --- \\

\midrule\\



ViTextVQA: A Large-Scale Visual Question Answering Dataset for Evaluating Vietnamese Text Comprehension in Images \citep{van2024vitextvqa} & N & N & No clear set of keywords but they mention having entities like person, sign, letter, window, shirt, tree, etc.\\

\midrule\\

HaVQA: A Dataset for Visual Question Answering and Multimodal Research in Hausa Language \citep{parida2023havqa} & N & N & Used Visual Genome dataset \citep{krishna2017visual}\\

\midrule\\



Navigating Text-to-Image Generative Bias across Indic Languages \citep{mittal2024navigating} &  N & N & 1000 images from COCO-NLLB  \\

\midrule\\

The World Wide Recipe: A community-centred framework for fine-grained data collection and regional bias operationalization \citep{magomere2024}  & Y & "food" & 765 dishes, with dish names collected in 131 local languages \\

\midrule\\

SEA-VQA: Southeast Asian Cultural Context Dataset For Visual Question Answering \citep{urailertprasert2024sea}  & N & N & Cambodia: Kun Lbokator, traditional martial arts in Cambodia; Lkhon Khol Wat Svay Andet \textit{and many more}. For more details, see table 4 in the paper. \\

\midrule\\

Extracting Cultural Commonsense Knowledge at Scale CCSK \citep{nguyen2023extracting}  & Y & "Food", "Drinks", "Clothing", "Rituals", "Traditions", "Behaviors", "Other" & --- \\

\midrule\\

Benchmarking Vision Language Models for Cultural Understanding \citep{nayak2024benchmarking} & Y & "traditions", "rituals", "food", "drink", "clothing" & --- \\

\midrule\\

Broaden the Vision: Geo-Diverse Visual Commonsense Reasoning \citep{yin2021broaden} &  N & N & only key words:wedding, family, religion, party, groom, bride, servant, couple, festival, school, student, pray, customer, funeral, dance, death, luck, children, soldier, chat, parents, meal, sports, restaurant, breakfast, muslim, sad, clothes, fast food, mother, happy, mourn, cook, women, wineglass, celebration, kitchen, audience, waiter, others. \\

\midrule\\

Towards Equitable Representation in Text-to-Image Synthesis Models with the Cross-Cultural Understanding Benchmark (CCUB) Dataset \citep{liu2023towards}  & Y & "architecture", "clothing", "city", "food and drink", "dance", "music", "visual arts", "nature", "people and action", "religion and festivals", "utensils and tools" & architecture (both interior and exterior), \\

\midrule\\

AltDiffusion: A Multilingual Text-to-Image Diffusion Model \citep{ye2024altdiffusion} &  Y & "Painting", "Literature", "Festival", "Food", "Clothes", "Landmark" & --- \\

\midrule\\

Partiality and Misconception: Investigating Cultural Representativeness in Text-to-Image Models \citep{10.1145/3613904.3642877} &  Y & "Architecture", "Artwork", "Festival activity", "Food", "Mythical figure", "Performance" & --- \\

\midrule\\

Multilingual Conceptual Coverage in Text-to-Image Models \cite{saxon-wang-2023-multilingual} & N & N & tangible nounseye, hand, head, smile, face, room, door, girl, person, man, love, watch, arm, hair, mother, car, mom, dad, table, phone, father, grin, mouth, kid, family, finger, world, shirt, ground, sister, chair, kitchen, woman, beer, hill, metal, hotel, princess, bench, detail, bird, cigarette, history, plastic, pizza, airplane, male, backpack, judge, dragon, sea, bike, female, garden, meal, toy, ship, flame, tail, library, weapon, cd, rope, cafeteria, porch, queen, duck, lake, television, boat, tent, roof, ticket, cop, milk, soldier, tank, thigh, belt, sandwich, bullet, teenager, apple, wine, supply, captain, cheese, feather, mask, prince, beaver, seal, stingray, shark, rose, bottle, mushroom, orange, pear, pepper, keyboard, lamp, telephone, couch, bee, beetle, butterfly, caterpillar, cockroach, tiger, wolf, bridge, castle, house, road, cloud, forest, mountain, camel, chimp, kangaroo, fox, raccoon, lobster, spider, worm, baby, crocodile, lizard, dinosaur, snake, turtle, hamster, rabbit, squirrel, tree, bicycle, train, tractor, jump, men, moon, clothes, neck, fire, tire, teacher, movie, dog, ring, eyebrow, sun, tall, doctor, sky, apartment, shoe, rock, daughter, girlfriend, bar, ball, hallway, tv, teeth, police, field, wife, brain, pants, tongue, cup, computer, bottom, bell, aunt, clock, suit, plate, chocolate, snow, guitar, truck, church, husband, van, blanket, bowl, mama, cookie, hat, monster, ceiling. \\

\midrule\\

Navigating Cultural Chasms: Exploring and Unlocking the Cultural POV of Text-To-Image Models \citep{ventura_chasms} & N & N & Moral Discipline and Social Values, Education, Economy, Religion, Health, Security, Aesthetics, Material Culture, Personality Characteristics and Emotions (adjective + ‘‘person’’), Social Capital Organizational Membership \\

\midrule\\

Beyond aesthetics: Cultural competence in text-to-image models - CUBE \citep{senthilkumar2024beyond}  & Y & "cuisine", "landmarks", "art" & --- \\

\midrule\\

Globally Inclusive Multimodal Multitask Cultural Knowledge Benchmarking \citep{schneider2025gimmickgloballyinclusive} &  Y & "Food", "Drinks", "Clothing", "Art", "Tools", "Sports", "Instruments", "Dance", "Music", "Rituals", "Traditions", "Festivals", "Customs", "Symbols", "Architecture", "Other" & "traditions", "rituals", "art", "music", "craftsmanship", "instruments", "festivals", "dance", "tools", "food", "clothing", "architecture", "sports", "location", "symbols", "drinks", "customs", "cultural significance", "theatre", "education", "culture", "games", "performing arts", "language", "performance", "characters", "practices", "skills", "origin", "cultural identity", "technology", "people", "community", "identity", "environment", "traditional medicine", "nature", "communication", "jewelry", "objects", "animal", "plants", "process", "agriculture", "celebrations", "details", "historical", "function or usage", "symbolism", "healthcare", "knowledge", "social status", "religion", "cultural space", "social space", "cultural practice", "unknown" \\

\midrule\\

WorldCuisines:A Massive-Scale Benchmark for Multilingual and Multicultural Visual Question Answering on Global Cuisines \citep{winata-etal-2025-worldcuisines}  & Y & "Food" & --- \\

\midrule\\

An image speaks a thousand words, but can everyone listen?On image transcreation for cultural relevance \citep{khanuja-etal-2024-image}  & Y & "Agriculture", "Beverages", "Birds", "Celebration", "Food", "Education", "Flower", "Clothing", "Fruit", "Houses", "Mammal", "Music", "Religion", "Sport", "Utensil", "Vegetable", "Visual Art" & --- \\

\midrule\\

GeoDE: a Geographically Diverse Evaluation Dataset for Object Recognition \citep{NEURIPS2023_d08b6801} & N & N & bag, chair, dustbin, hairbrush/comb, hand soap, hat, light fixture, light switch, toothbrush, toothpaste/toothpowder, candle, cleaning equipment, cooking pot, jug, lighter, medicine, plate of food, spices, stove, toy,backyard, car, fence, front door, house, road sign, streetlight/lantern, tree, truck, waste container, bicycle, boat, bus, dog, flag, monument, religious building, stall, storefront, wheelbarrow \\

\midrule\\

CIC: A framework for Culturally-aware Image Captioning \citep{youngsik_cic} & Y & "architecture", "clothing", "dance and music", "food and drink", "religion" & --- \\

\midrule\\



ViSAGe: A Global-Scale Analysis of Visual Stereotypes in Text-to-Image Generation \citep{jha-etal-2023-seegull} & N & N & 385 out of the original 1994 attributes, where these attributes are from SeeGULL \citep{jha-etal-2023-seegull}\\

\midrule\\

DIG In: Evaluating Disparities in Image Generations with Indicators for Geographic Diversity \citep{hall2023dig} & N & N & Please see section 3.1, they use real world datasets \\

\midrule\\

Exploiting Cultural Biases via Homoglyphs in Text-to-Image Synthesis \citep{struppek_exploiting}  & Y & "People", "Buildings", "Misc" & See Appendix B.1 of the paper\\

\midrule\\

Inspecting the Geographical Representativeness of Images from Text-to-Image Models \citep{10376690}  & Y & city, beach, house, festival, road, dress, flag, park, wedding, and kitchen & --- \\

\midrule\\

Easily Accessible Text-to-Image Generation Amplifies Demographic Stereotypes at Large Scale \citep{bianchi_easily} &  N & --- & race, gender, ethnicity, and occupation \\

\midrule\\

Cultural Concept Adaptation on Multimodal Reasoning \citep{li-zhang-2023-cultural} & Y & "festival", "music", "religion and belief", "animal and plant", "food", "clothing", "building", "agriculture", "tool", "sport" & --- \\

\midrule\\

CROPE: Evaluating In-Context Adaptation of Vision and Language Models to Culture-Specific Concepts \citep{nikandrou-etal-2025-crope} & Y & "animal", "food and beverages", "clothing and grooming", "the house", "agriculture and vegetation", "basics and technology", "motion", "time", "cognition", "speech and language","religion and belief" & MarVL and MMC  \\
\midrule\\

Exploring Visual Culture Awareness in GPT-4V: A Comprehensive Probing \citep{cao2024exploring} & Y & "animal", "food and beverages", "clothing and grooming", "the house", "agriculture and vegetation", "basics and technology", "motion", "time", "cognition", "speech and language","religion and belief" & --- \\

\midrule\\

RAVENEA: A Benchmark for Multimodal Retrieval-Augmented Visual Culture Understanding \citep{li2025ravenea} & Y & "Architecture", "Cuisine", "History", "Art", "Daily Life", "Companies", "Sports \& Recreation", "Transportation", "Religion", "Nature", "Tools" & --- \\


\bottomrule
\end{longtable}